\documentclass[conference]{IEEEtran}
\IEEEoverridecommandlockouts
\pdfoutput=1
\usepackage{cite}
\usepackage{amsmath,amssymb,amsfonts}
\usepackage{algorithmic}
\usepackage{graphicx}
\usepackage{textcomp}

\usepackage{times}  
\usepackage{helvet}  
\usepackage{courier}  
\usepackage{url}  
\usepackage{graphicx}  
\usepackage{multirow}
\usepackage{array}
\usepackage{longtable}
\usepackage{rotating}
\usepackage{diagbox}
\usepackage{supertabular}
\usepackage{amssymb}
\usepackage{pifont}
\usepackage{float}
\usepackage{amsmath}
\usepackage{subfigure}
\usepackage{booktabs}
\usepackage{caption}

\usepackage[table,xcdraw]{xcolor}
\usepackage{array}
\newtheorem{myDef}{Definition}
\newcommand{\tabincell}[2]{\begin{tabular}{@{}#1@{}}#2\end{tabular}}

\def\BibTeX{{\rm B\kern-.05em{\sc i\kern-.025em b}\kern-.08em
    T\kern-.1667em\lower.7ex\hbox{E}\kern-.125emX}}
\begin{document}

\title{CNNTOP: a CNN-based Trajectory Owner Prediction Method}

\author{\IEEEauthorblockN{Xucheng Luo, Shengyang Li, Yuxiang Peng}
\IEEEauthorblockA{\textit{School of Information and Software Engineering, University of Electronic Science and Technology of China} \\
Chengdu 610054, China\\
xucheng@uestc.edu.cn}
}

\maketitle

\begin{abstract}
Trajectory owner prediction is the basis for many applications such as personalized recommendation, urban planning. Although much effort has been put on this topic, the results archived are still not good enough. Existing methods mainly employ RNNs to model trajectories semantically due to the inherent sequential attribute of trajectories. However, these approaches are weak at Point of Interest (POI) representation learning and trajectory feature detection. Thus, the performance of existing solutions is far from the requirements of practical applications. In this paper, we propose a novel CNN-based Trajectory Owner Prediction (CNNTOP) method. Firstly, we connect all POI according to trajectories from all users. The result is a connected graph that can be used to generate more informative POI sequences than other approaches. Secondly, we employ the Node2Vec algorithm to encode each POI into a low-dimensional real value vector. Then, we transform each trajectory into a fixed-dimensional matrix, which is similar to an image. Finally, a CNN is designed to detect features and predict the owner of a given trajectory. The CNN can extract informative features from the matrix representations of trajectories by convolutional operations, Batch normalization, and $K$-max pooling operations. Extensive experiments on real datasets demonstrate that CNNTOP substantially outperforms existing solutions in terms of macro-Precision, macro-Recall, macro-F1, and accuracy.
\end{abstract}


\section{Introduction}
\label{sec:introduction}
Internet applications such as Foursquare, Twitter, Google Places, Instagram, and Airbnb, provide Location-based Social Network (LBSN) data, which includes social network among users and/or check-in locations. Such information enables human mobility understanding~\cite{Yang2019,Zhou2019,Gao2019,Cho2011} and is crucial for applications like personalized recommendation, intelligent transportation systems, urban planning~\cite{Marti2019,Zhang2016}, and so forth.

A critical task of human mobility understanding is to predict the owner of a given trajectory, which is defined as the pathway followed by certain users. Determination of the owner-trajectories relationship greatly benefits many real-world applications through performance enhancement for personalized recommendation and urban plan improvement for traffic mitigation. Bike or car-sharing APPs, like Mobike and Uber, collected a large amount of trajectory data. For a specific application, the owner of a given trajectory typically corresponds to the owner of the account of the APP. However, this general understanding is invalid if the account is shared by multiple users. Users of multiple trajectory-collecting APPs may also sign in different APPs with various usernames. The trajectory-user relationship, therefore, cannot be clearly defined when data from multiple APPs are analyzed in aggregation without further manipulation.

The trajectory owner prediction is essentially determined by trajectory models. Existing modeling methods for human trajectories mainly include Markov Chains (MC) and Recurrent Neural Networks (RNN). MC-based approaches~\cite{Zhang2016} assume strong independence among non-adjacent locations. Thus they cannot capture the long term dependency of locations in a trajectory. Generally, RNNs are thought to be the ideal candidate for modeling trajectories since they can take variable-length trajectories as input and capture long term dependency of locations. TULER~\cite{Gao2017} is the first method that leverages RNN to address trajectory and user linking. TULVAE~\cite{Zhou2018} improves TULER by combining RNN and VAE to address data sparsity, structural dependency, and shallow generation. Although the above methods are suitable for capturing geographic and semantic features for some tasks such as next location prediction~\cite{Feng2018}, their performance at capturing features for trajectory owner prediction is far from being satisfactory.

In summary, existing approaches have two weaknesses. On the one hand, they ignore the fact that the owner of a trajectory is mainly determined by some key POIs or POI tuples such as apartments, working places, and leisure venues. Therefore, existing approaches treat trajectories as sequences and leverage RNN to learn trajectory representations. However, it is reported that RNN-based approaches are good at capture long term semantics but weak at feature detection compared to Convolutional Neural Networks (CNNs) in Natural Language Processing (NLP)~\cite{Yin2016}. On the other hand, they employ word2vec algorithm to generate POI embeddings directly from trajectories, which leads to low-quality POI embeddings due to inadequate POI sequence coverage. Therefore, the trajectory representation which depends on POI embeddings is not efficient enough to discriminate its owner.

In this paper, we propose a novel CNN-based Trajectory Owner Prediction (CNNTOP) to overcome the above weaknesses. CNNTOP represents trajectories with matrices and leverages CNN to efficiently detect features to predict the owner of a given trajectory. The main contributions of this work are as follows:
\begin{enumerate}
	\item CNNTOP connects all trajectories to form a linked graph and employs the Node2Vec algorithm to encode each POI into a low-dimensional real value vector. Compared with existing designs, our scheme can generate more effective POI sequences and POI representations.
	\item CNNTOP represents trajectory segmentations with matrices and designs a Convolutional Neural Network to detect trajectory features from corresponding matrices. Compared with existing solutions, our scheme can efficiently extract trajectory features for predicting corresponding owners.
	\item Extensive experiments are conducted on three real-world datasets. The performances of different methods are compared. Moreover, parameter sensitivity is also analyzed.
\end{enumerate}

The remainder of this paper is organized as follows. In Section~\ref{Relatedwork}, we present the existing work related to trajectory owner prediction. After that, the problem statement is given in Section~\ref{problem}. Section~\ref{Method} describes our design for efficient trajectory owner prediction. Section~\ref{Exp} presents the evaluation method and experimental results. We summarize the paper and outline the direction for future work in Section~\ref{Conclusion}.

\section{Related work}
\label{Relatedwork}

In this section, we review some related work on owner prediction for trajectories, including sequential methods based on Markov Chains and RNN-based methods. Then, we introduce the convolutional neural network which we employ to solve the TOP problem in this work.

\textbf{Human Trajectory Representation.} Trajectory Owner Prediction (TOP) is a critical aspect of human mobility understanding, which has recently attracted dense interests from academia and industry~\cite{Cheng2013,Zheng2015,Zhang2016,Zhuang2017}. The performance of the TOP problem depends on trajectory representation. Markov chain-based approach~\cite{Qiao2018} employs Markov chain and other tricks to predict human mobility. This kind of method cannot take into account long term dependency due to the Markov property. TULER~\cite{Gao2017} adopts word2vec algorithm~\cite{Mikolov2013a} to learn low-dimensional vector representations of POIs and employs RNN to encode a given trajectory to a real-value vector. A supervised classifier is designed in TULER to map vector representations of trajectories to users. DeepMove~\cite{Feng2018} and DPLink~\cite{Feng2019} also leverage RNN to learn trajectory representations. TULVAE~\cite{Zhou2018} is an improved version of TULER. It incorporates a VAE model in the RNN seq2seq pipeline to mitigate data sparsity, structural dependency, and shallow generation. Generally speaking, RNN is very suitable for modeling sequential data since it can take variant length sequence as input and output a fixed-length vector. However, the performance of modeling trajectories with RNN depends on specific tasks. As to the TOP problem, the performance of existing RNN-based methods is far from real applications.

\textbf{Convolutional Neural Network (CNN).} CNN is a kind of deep neural network, which is characterized by its convolutional layers and pooling layers. Typically, the input layer is followed by a convolutional layer and then a pooling layer. After that, there may be multiple convolutional layer and pooling layer combinations. Then, a fully connected layer is followed. LeNet~\cite{Lecun1998} is a classic instance of CNN. The convolutional layers are used for feature detection. The pooling layers are used for selecting main features and preventing overfitting. It can also keep invariant such as translation and rotation. CNN is typically used for detecting meaningful patterns in images, for example, facial expression recognition~\cite{Kuo2018}, object detection~\cite{Ren2017}. Recently, CNN has also been used to process sequential data, for example, Neural Machine Translation~\cite{Gehring2017}, text classification~\cite{Kim2014}. Extensive experiments have demonstrated that the performance of RNN or CNN for sequential data modeling depends on how important it is to understand the whole sequence semantically~\cite{Yin2017}. For many NLP tasks such as modeling sentence pairs~\cite{Yin2016}, sentence classification~\cite{Kim2014}, CNN outperforms RNN. Therefore, it is promising to solve the TOP problem with CNN.

In summary, existing solutions mainly treat trajectories as time series and model them with Markov chains or RNNs. However, trajectory owner prediction is more sensitive to feature detection. Thus we propose a novel CNN-based approach to address the TOP problem. As far as we know, our work is the first one that employs CNN for trajectory owner prediction.

\section{Problem statement}
\label{problem}
We consider the trajectory owner prediction problem as follows. With an LBSN system, there are many users and POIs. The set of users can be denoted as $U=\{u_1,u_2, \dots, u_N\}$, where $N$ is the number of users. The set of POIs is denoted by $L=\{l_1, l_2, \dots, l_m\}$, where $m$ is the number of POIs. Users report their check-ins sequentially. Consequently, each user has a long POI trajectory. Let $T_{u_i}=<l_{i1},l_{i2},\dots,l_{in}>$ denote a trajectory generated by the user $u_i$ during a time interval, where $l_{ij}\in L$ is a POI. Since there are many users in an LBSN system, a lot of trajectories are generated in a given period. Furthermore, long trajectories can be subdivided into many segments. Therefore, we have a set of trajectories $T=\{t_1, t_2, \dots, t_M\}$, where $M >> N$. For clarity, we present two example trajectories below.
\begin{equation*}\label{example-tr}
\begin{gathered}
  {\text{8 622 474 474 474 481 482 482 83}} \hfill \\
  {\text{8 83   270 487 270 270   83   83 471}} \hfill \\
\end{gathered}
\end{equation*}
The first number, i.e. 8, is the user ID. Following that are the POI IDs which user 8 has been. Thus, both the above two trajectories are generated by user 8.

Based on above settings, The trajectory owner prediction problem can be formally defined as below.
\begin{myDef}{Trajectory owner prediction problem.}
Suppose we are given 1) a user set $U=\{u_1,u_2, \dots, u_N\}$, 2) a POI set $L=\{l_1, l_2, \dots, l_m\}$, 3) a trajectory set $T=\{t_1, t_2, \dots, t_M\}$, and 4) a subset $T_s\subset T$ and the owner of each trajectory in $T_s$ is known. Based on above inputs, trajectory owner prediction aims to output the owner of each trajectory in $T\setminus T_s$.
\end{myDef}

\section{Methodology}
\label{Method}

This section describes the trajectory owner prediction methodology. We first present the architecture of CNNTOP. Then, we describe the POI representation learning. After that, the prediction algorithm is put forth. Finally, the model training is depicted.
\begin{figure}
  \centering
  \includegraphics[scale=0.48]{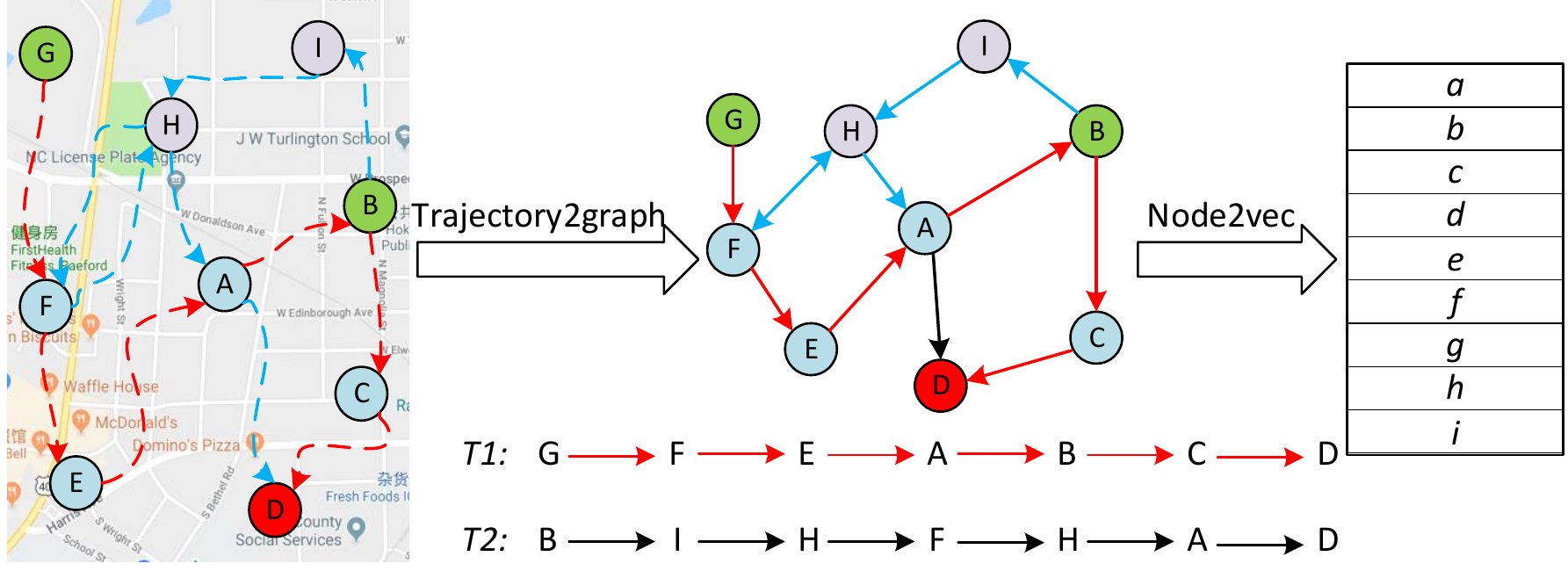}
  \caption{The process of POI embedding. Firstly, trajectories are connected to form a graph. Then, Node2Vec algorithm is applied to embed POIs to low-dimensional space.}
  \label{fig:cnntop-n2v}
\end{figure}

\begin{figure}
  \centering
  \includegraphics[width=0.5\textwidth]{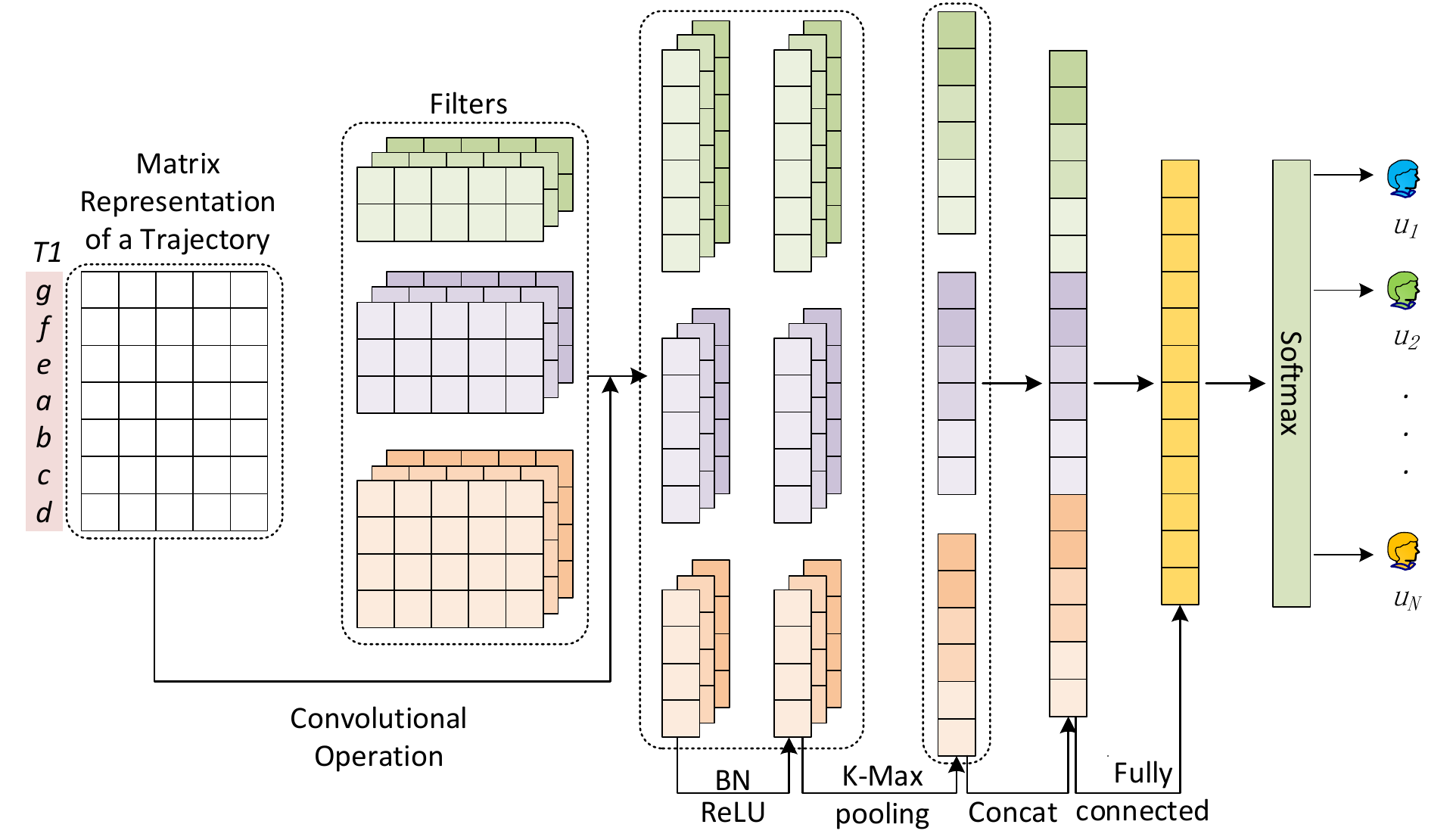}
  \caption{The CNN-based trajectory classification. The matrix representations of trajectories are constructed according to the POI embeddings and check-in order in trajectories. A convolutional neural network is employed to extract features of a trajectory and predict its owner.}
  \label{fig:cnntop-class}
\end{figure}
\subsection{CNNTOP Architecture}
Trajectory owner prediction mainly depends on the trajectory representation which is in turn determined by the POI embeddings. Since users and trajectories are typically linked by some key POIs or POI tuples, which is very similar to $n$-gram features of texts, algorithms that are good at detecting $n$-gram features are preferred here. Inspired by that CNNs are good at $n$-gram feature detection~\cite{Yin2016}, we employ 1-D CNN to extract features of trajectories. As to POI embedding, which should fully reflect the geographical property, we consider network embedding algorithm to address this problem. Therefore, CNNTOP is comprised of two parts. The first part is POI representation learning which is illustrated in Fig.\ref{fig:cnntop-n2v}. The second part is CNN-based trajectory owner prediction which is depicted in Fig.\ref{fig:cnntop-class}. We present the details of the two parts as follows.

\subsection{POI Representation Learning}
POI representation learning maps POIs to low-dimensional vectors, which is crucial for trajectory owner prediction. Good POI representations lead to good trajectory representations and thus achieve high performance for owner prediction. Generally, POI representation should incorporate enough geographical features to reflect the mutual relation among POIs. Thus, the key POIs or POI tuples could be detected to predict the owner of a given trajectory. For each trajectory, it is obviously a sequence. Therefore, existing approaches treat trajectories as sentences and encode each POI by utilizing word2vec~\cite{Mikolov2013a} algorithm. Different from that, we first construct a graph $G$. The nodes of the graph $G$ are all the POIs. If there is a user who moved from $l_i$ to $l_j$, there is an edge from $l_i$ to $l_j$. In this way, we can construct a linked graph. This process is dubbed Trajectory2Graph in the Fig.\ref{fig:cnntop-n2v}.

Then, we employ Node2Vec~\cite{Grover2016} algorithm to embed each POI into a low-dimensional vector. Essentially, Node2Vec leverages word2vec~\cite{Mikolov2013a} algorithm, which is an algorithm for learning a vector representation of a word based on the context in a large corpus, to embed nodes to low-dimensional real-value vectors. However, the key difference is that Node2Vec employs random walks on a graph to generate sequences while word2vec directly takes use of existing sequences. Therefore, the former can produce more sequences. Specifically, Node2Vec performs a biased random walk to balance depth-first sampling and breadth-first sampling, which control walks by two parameters $p$ and $q$ to capture structural equivalence and homophily. The flexibility of Node2Vec in exploring neighborhoods enhances the node representation learning. After several rounds of random walks, we obtain a set of node sequences of $G$ and then send them to the word2vec algorithm. Let $F:L \to {R^S}$ be the function which maps nodes to low-dimensional vectors. Here $S$ is the dimensionality of the embedding space ($S$ generally ranges from tens to hundreds). To embed POIs to low-dimensional vectors, the objective function of Node2Vec is defined below:
\begin{equation}
	\label{objective-function}
	\mathop {\max }\limits_F \sum\limits_{l \in L} {[ - \log {Z_l}}  + \sum\limits_{{l_i} \in N(l)} {F({l_i})}  \cdot F(l)],
\end{equation}
where ${Z_l} = \sum\nolimits_{c \in L} {\exp (F(l) \cdot F(c))}$, $N(l) \subset L$ is the network neighborhood of node $l$ generated through random walk strategy. Finally, we learn a vector $V_{l}$ corresponding to each POI $l \in L$ in an $S$ dimensional space. The space complexity of the algorithm is $O({d^2}|L|)$ where $d$ is the average degree of the graph $G$.

Compared with existing solutions, the POI representation learning method of CNNTOP has two advantages. Firstly, more POI sequences can be generated. For existing solutions, the POI sequences for POI representation learning is from trajectories, for example, $<l_1, l_2, l_3, l_4,l_5>$, $<l_5, l_6, l_7, l_8, l_9>$. However, there are no POI sequences cross the above two samples, for instance, $<l_3, l_4, l_5, l_6, l_7>$, and $<l_4, l_5, l_6, l_7, l_8>$. Since CNNTOP construct a connected graph, all the aforementioned sequences can be generated. Secondly, more informative POI sequences can be generated. Since Node2vec employs biased random walks to generate POI sequences. Appropriate control of depth-first sampling and breadth-first sampling can yield sequences that embody sufficient structural information. Therefore, the sequences generated in this way can lead to more efficient POI representations.

\subsection{Trajectory Owner Prediction}
Based on the vector representations of POIs, we can construct matrix representations of trajectories. For each trajectory, we arrange corresponding POI vectors as rows of a matrix sequentially. In this approach, we obtain the matrix representation of a trajectory. Now that the trajectories are represented as matrices, we can treat each trajectory as an image. Therefore, we can employ CNN to process each trajectory. We designed a five-layer CNN to do this, which is depicted in Fig.\ref{fig:cnntop-class}.

The first layer is the input layer which is an $h\times embedding\_ size$ matrix $X$, where $embedding\_ size$ is the same as the length of node vectors. The $h$ represents the length of a given trajectory segmentation and can be fixed or variable. If $h$ is fixed, it is very convenient for model training with mini-batch SGD. In order to have $h$ fixed, for some cases where the lengths of trajectories are less than $h$, we append one or more same padding vectors to the end of them. The elements of the padding vector are averages of corresponding dimensions of all vector representations of POIs. If $h$ is variable, the trajectory matrix can represent a variable-length trajectory without padding. In this case, we can only employ SGD algorithm to train our model, which is time-consuming.

The second layer is the convolutional layer. CNNTOP adopts three kinds of convolutional kernels which are $m \times embedding\_size$. The $m$ is configured as 2, 3, and 4 respectively. The $embedding\_size$ is also the same as the length of node vectors. The number of each kind of convolutional kernel is 64. To prevent the ``gradient disappear" problem, the activation function of this layer is the Rectified Linear Unit (ReLU). Essentially, CNNTOP employs 1-D convolution to detect $n$-gram features in trajectories. In short, the output of this layer is shown below.
\begin{equation}\label{eq:conv}
  O_{conv}=ReLU(CONV(W_{conv},X) + b_{conv}),
\end{equation}
where $CONV$, $W_{conv}$, and $b_{conv}$ are convolutional operation, filters and bias, respectively.

The third layer is the Batch Normalization (BN)~\cite{Ioffe2015} layer, which is used to normalize the output of the convolutional layer by adjusting and scaling the activations of the convolutional layer. The BN operation can improve the speed, performance, and stability of CNNTOP. The output of this layer is denoted by $BN(O_{conv})$, where $BN$ is Batch Normalization operation.

The fourth layer is the pooling layer. We adopt $k$-max pooling to preserve the top $k$ important values from the output of each Batch Normalized convolutional operation. In this way, CNNTOP can extract the most important features while keeps the outputs of pooling operation for different trajectories in the same size. The output of the pooling layer is depicted as formula \ref{eq:convx}.
\begin{equation}
\label{eq:convx}
C_i= P(BN(O_{conv})),
\end{equation}
where $P$ is the $k$-max pooling operation.

After the pooling layer, we obtain vector representation of a trajectory by concatenation of $C_i$. The result is denoted by $C=[C_1\colon C_2\colon \dots\colon C_n]$, where $n$ is the number of convolutional kernels. Then, a fully connected layer is followed. The output of the fully connected layer is the input of the softmax function. Through it, the probability distribution of different categories is calculated as formula \ref{eq:prob}.
\begin{equation}\label{eq:prob}
\hat{y} = softmax(WC + B),
\end{equation}
where $W$ and $B$ are weights and bias of the fully connected layer, respectively. The output $\hat{y}\in R^N$ is the prediction probability distribution, where $N$ is the number of users or labels.

For each trajectory, the loss function is defined as the cross-entropy of the prediction and the ground truth. It can be calculated as below:
\begin{equation}\label{eq:loss}
  L(\hat{y}) = -\sum_{i=1}^{N}y_ilog(\hat{y}_i),
\end{equation}
where $y_i$ is the $i$-th element of $y$ which is the one-hot encoding vector of the ground truth owner.

\subsection{Model Training}

To accelerate the training process, we use Adam algorithm~\cite{Kingma2015} to optimize the objective function which is shown below.
\begin{equation}
\label{eq:costfunc}
 min-\frac{1}{B}\sum_{j=1}^{B}\sum_{i=1}^{m}y_{ji}log(\hat{y}_{ji}),
\end{equation}
where $B$ is the number of instances in the current batch of trajectories. The mini-batch training of a CNN model requires the same size inputs. However, trajectories typically have different lengths. To tackle this problem, we pad each trajectory to the same size. If the length of the longest trajectory is $l$, the dimension of the matrix representations of all trajectories should be $l\times embedding\_size$. For trajectories with length $s (s < l)$, we append $l-s$ padding vectors to the end of them. The elements of the padding vector are the averages of corresponding elements of all POI vectors.

\section{Evaluation}
\label{Exp}
In this section, we first describe the datasets. Then, we introduce the baselines and performance metrics, respectively. Finally, we present the experimental results.

\subsection{The Datasets}
We conduct our experiments on three publicly available datasets: Brightkite\footnote{http://snap.stanford.edu/data/loc-brightkite.html}, Foursquare\footnote{https://sites.google.com/site/yangdingqi/home/foursquare-dataset}, and Gowalla\footnote{http://snap.stanford.edu/data/loc-gowalla.html}. All these datasets contain users and their check-in sequences. The statistics of the original datasets are described in Table\ref{tab:dataset}. To perform experiments, we further process the original datasets. For each user, we subdivide his/her check-in sequences into sub-sequences (trajectories) based on the given time interval which we define to be six hours to reflect the periodicity of human activity. The label of each trajectory is its owner.

\begin{table}
\centering
\caption{Summary statistics of datasets.}
\label{tab:dataset}
\arrayrulecolor{black}
\begin{tabular}{|c|c|c|c|c|c|}
\hline
\rule{0pt}{10pt}Dataset    & \#Users & \#Trajectories & \#POIs & \tabincell{c}{Average\\ Length} & Range         \\
\hline
\multirow{2}{*}{BrightKite} & \rule{0pt}{10pt}100      & 7089    & 757    & 143.14      & {[}1, 139]   \\
\cline{2-6}
                            & \rule{0pt}{10pt}200      & 15501   & 1602  & 299         & {[}1, 116]   \\
\hline
\multirow{2}{*}{Foursquare} & \rule{0pt}{10pt}100      & 13079   & 6443   & 238         & {[}1, 24]     \\
\cline{2-6}
                            & \rule{0pt}{10pt}200      & 24175   & 11050  & 216         & {[}1,55]      \\
\hline
\multirow{2}{*}{Gowalla}    & \rule{0pt}{10pt}100      & 3683     & 2283   & 57          & {[}1, 29]    \\
\cline{2-6}
                            & \rule{0pt}{10pt}200      & 7421     & 4667   & 58          & {[}1, 55]    \\
\hline

\end{tabular}
\end{table}

\subsection{Baselines}
There are many methods addressing trajectory owner prediction. Among them, TULER\cite{Gao2017} and TULVAE\cite{Zhou2018} obtained the best performance so far. Therefore we mainly compare our method with them.

\textbf{TULER} predicts the trajectory owner via trajectory embedding. First, TULER employs the word2vec algorithm to embed POIs to low-dimensional vectors. Then, an RNN with Gated Recurrent Unit (GRU) or LSTM is used to transform a trajectory into a low-dimensional semantic vector. The output layer of the RNN is softmax which outputs the probability distribution of users for a given trajectory. Although TULER has several variants, including TULER-LSTM, TULER-GRU, TULER-LSTM-S, TULER-GRU-S, Bi-TULER, HTULER-L, HTULER-G, and HTULER-B, we only compare the performance with Bi-TULER since it has the best performance among all the variants.

\textbf{TULVAE} can be treated as an improved version of TULER. It is a variant of the seq2seq model. It first employs RNN to encode a trajectory into a vector. Then, a VAE is followed. After that, an RNN decoder is connected. In this way, the latent representation of a trajectory can be further bounded. Therefore, TULVAE mitigates the problem of data sparsity and obtain better performance than that of TULER.

\subsection{Performance Metrics}

Essentially, the trajectory owner prediction is a multiclass classification problem. Therefore, we will use average accuracy ($ACC@k$), macro-Precision ($macro\text{-}P$), macro-Recall ($macro\text{-}R$), and  macro F1-score ($macro\text{-}F1$) to evaluate our method. The $ACC@k$ metric is formulated as:
\begin{equation}
\label{eq:acc}
ACC@k = \frac{\#\text{correctly predicted trajectories}@k}{\#\text{trajectories}}
\end{equation}

To present the details of the remainder metrics, we need to define some metrics for a specific class. For a specific class label $C_i$, we treat $C_i$ as a positive class, but all other classes as negative ones. Thus we have the following metrics.
\begin{itemize}
  \item $TP_i$: the number of instances which the algorithm outputs is positive while the ground truth is also positive.
  \item $TN_i$: the number of instances which the algorithm outcome is negative while the ground truth is also negative.
  \item $FP_i$: the number of instances which the algorithm outputs is positive whereas the ground truth is negative.
  \item $FN_i$: the number of instances which the algorithm outputs is negative whereas the ground truth is positive.
\end{itemize}

Based on the above metrics, the calculation of the aforementioned performance metrics are as follows:
\begin{equation}
\label{eq:precision}
  macro\text{-}P = \frac{\sum_{i=1}^{l}\frac{TP_i}{TP_i+FP_i}}{l}
\end{equation}

\begin{equation}
\label{eq:recall}
  macro\text{-}R = \frac{\sum_{i=1}^{l}\frac{TP_i}{TP_i+FN_i}}{l}
\end{equation}
The $macro\text{-}F1$ is the harmonic mean of $macro\text{-}P$ and $macro\text{-}R$, which is depicted as below.
\begin{equation}
\label{eq:macro-f1}
 macro\text{-}F1 = 2*\frac{macro\text{-}P*macro\text{-}R}{macro\text{-}P + macro\text{-}R}
\end{equation}

\begin{table*}
\centering
\caption{Performance comparison of different methods.}
\label{tab:results}
\arrayrulecolor{black}
\begin{tabular}{!{\color{black}\vrule}l!{\color{black}\vrule}l!{\color{black}\vrule}l!{\color{black}\vrule}l!{\color{black}\vrule}l!{\color{black}\vrule}l!{\color{black}\vrule}l!{\color{black}\vrule}l!{\color{black}\vrule}l!{\color{black}\vrule}l!{\color{black}\vrule}l!{\color{black}\vrule}l!{\color{black}\vrule}}
\hline
\multirow{2}{*}{Dataset}         & \multirow{2}{*}{Method} &\rule{0pt}{10pt} ACC@1   & ACC@5   & macro-P & macro-R & macro-F1      & ACC@1   & ACC@5   & macro-P & macro-R & macro-F1       \\
\cline{3-12}
                                 &\rule{0pt}{10pt}                         & \multicolumn{5}{c!{\color{black}\vrule}}{Trajectory Length = 20} & \multicolumn{5}{c!{\color{black}\vrule}}{Trajectory Length = 30}  \\
\hline
\multirow{4}{*}{\rotatebox{90}{\tabincell{c}{BrightKite\\(100)}}} & \rule{0pt}{10pt}Bi-TULER                   & 49.8\%  & 61.3\%  & 21.10\% & 20.38\% & 20.74\%       & 51.00\% & 63.12\% & 21.23\% & 21.01\% & 21.12\%        \\
\cline{2-12}
                                 & \rule{0pt}{10pt}TULVAE                  & 63.8\%  & 71.93\% & 30.05\% & 29.20\% & 29.62\%       & 62.84\% & 73.01\% & 31.40\% & 28.07\% & 29.77\%        \\
\cline{2-12}
                                 & \rule{0pt}{10pt}CNNTOP                  & \textbf{92.19}\% & \textbf{95.95}\% & \textbf{90.67}\% & \textbf{90.79}\% & \textbf{90.73}\%       & \textbf{91.97}\% & \textbf{95.81}\% & \textbf{90.57}\% & \textbf{90.60}\% & \textbf{90.59}\%        \\
\cline{2-12}
                                 & \rule{0pt}{10pt}improvement                & 28.33\% & 24.02\% & 60.62\% & 61.59\% & 61.11\%       & 29.13\% & 22.80\% & 59.17\% & 62.53\% & 60.82\%        \\
\hline
\multirow{4}{*}{\rotatebox{90}{\tabincell{c}{BrightKite\\(200)}}} & \rule{0pt}{10pt}Bi-TULER                   & 51.34\% & 64.22\% & 25.55\% & 21.23\% & 23.19\%       & 50.18\% & 60.19\% & 25.31\% & 22.64\% & 23.90\%        \\
\cline{2-12}
                                 & \rule{0pt}{10pt}TULVAE                  & 57.82\% & 67.07\% & 34.82\% & 29.87\% & 32.16\%       & 56.90\% & 67.10\% & 34.48\% & 30.18\% & 32.19\%        \\
\cline{2-12}
                                 &\rule{0pt}{10pt} CNNTOP                  & \textbf{91.60}\% & \textbf{95.35}\% & \textbf{90.61}\% & \textbf{91.32}\% & \textbf{90.97}\%       & \textbf{91.63}\% & \textbf{95.34}\% & \textbf{89.96}\% & \textbf{90.15}\% & \textbf{90.06}\%        \\
\cline{2-12}
                                 & \rule{0pt}{10pt}improvement                & 33.78\% & 28.28\% & 55.79\% & 61.45\% & 58.81\%       & 34.73\% & 28.24\% & 55.48\% & 59.97\% & 57.87\%        \\
\hline
\multirow{4}{*}{\rotatebox{90}{\tabincell{c}{Foursquare\\(100)}}} & \rule{0pt}{10pt}Bi-TULER                   & 13.03\% & 28.15\% & 15.72\% & 8.89\%  & 11.36\%       & 16.79\% & 26.86\% & 17.59\% & 10.19\% & 12.91\%        \\
\cline{2-12}
                                 & \rule{0pt}{10pt}TULVAE                  & 45.37\% & 54.50\% & 48.73\% & 38.72\% & 43.15\%       & 45.85\% & 54.68\% & 47.12\% & 39.16\% & 42.77\%        \\
\cline{2-12}
                                 & \rule{0pt}{10pt}CNNTOP                  & \textbf{64.34}\% & \textbf{72.73}\% & \textbf{64.94}\% & \textbf{63.91}\% & \textbf{64.42}\%       & \textbf{64.84}\% & \textbf{73.71}\% & \textbf{65.10}\% & \textbf{64.16}\% & \textbf{64.63}\%        \\
\cline{2-12}
                                 & \rule{0pt}{10pt}improvement                & 18.97\% & 18.23\% & 16.21\% & 25.19\% & 21.27\%       & 18.99\% & 19.03\% & 17.98\% & 25.00\% & 21.86\%        \\
\hline
\multirow{4}{*}{\rotatebox{90}{\tabincell{c}{Foursquare\\(200)}}} & \rule{0pt}{10pt}Bi-TULER                   & 7.62\%  & 13.14\% & 8.06\%  & 5.18\%  & 6.31\%        & 6.99\%  & 13.03\% & 8.24\%  & 4.76\%  & 6.04\%         \\
\cline{2-12}
                                 & \rule{0pt}{10pt}TULVAE                  & 32.38\% & 40.32\% & 34.13\% & 27.39\% & 30.39\%       & 31.95\% & 40.49\% & 33.40\% & 26.92\% & 29.81\%        \\
\cline{2-12}
                                 &\rule{0pt}{10pt} CNNTOP                  & \textbf{56.22}\% & \textbf{64.85}\% & \textbf{56.94}\% & \textbf{56.80}\% & \textbf{56.87}\%       & \textbf{56.40}\% & \textbf{64.52}\% & \textbf{57.11}\% & \textbf{56.88}\% & \textbf{56.99}\%        \\
\cline{2-12}
                                 &\rule{0pt}{10pt} improvement                & 23.84\% & 24.53\% & 22.81\% & 29.41\% & 26.48\%       & 24.45\% & 24.03\% & 23.71\% & 29.96\% & 27.18\%        \\
\hline
\multirow{4}{*}{\rotatebox{90}{\tabincell{c}{Gowalla\\(100)}}}    & \rule{0pt}{10pt}Bi-TULER                   & 23.68\% & 36.04\% & 8.30\%  & 6.74\%  & 7.44\%        & 26.25\% & 38.35\% & 9.09\%  & 8.07\%  & 8.55\%         \\
\cline{2-12}
                                 & \rule{0pt}{10pt}TULVAE                  & 46.66\% & 59.38\% & 37.55\% & 25.05\% & 30.06\%       & 47.19\% & 59.05\% & 35.74\% & 25.36\% & 29.67\%        \\
\cline{2-12}
                                 & \rule{0pt}{10pt}CNNTOP                  & \textbf{65.34}\% & \textbf{74.57}\% & \textbf{73.84}\% & \textbf{71.87}\% & \textbf{72.84}\%       & \textbf{66.90}\% & \textbf{74.57}\% & \textbf{74.14}\% & \textbf{72.58}\% & \textbf{73.35}\%        \\
\cline{2-12}
                                 & \rule{0pt}{10pt}improvement                & 18.68\% & 15.19\% & 36.29\% & 46.82\% & 42.78\%       & 19.71\% & 15.52\% & 38.40\% & 47.22\% & 43.68\%        \\
\hline
\multirow{4}{*}{\rotatebox{90}{\tabincell{c}{Gowalla\\(200)}}}    & \rule{0pt}{10pt}Bi-TULER                   & 11.51\% & 25.13\% & 6.86\%  & 4.70\%  & 5.58\%        & 12.40\% & 21.54\% & 6.93\%  & 5.03\%  & 5.83\%         \\
\cline{2-12}
                                 & \rule{0pt}{10pt}TULVAE                  & 46.77\% & 54.65\% & 44.87\% & 33.81\% & 38.57\%       & 47.04\% & 55.10\% & 45.90\% & 34.14\% & 39.16\%        \\
\cline{2-12}
                                 & \rule{0pt}{10pt}CNNTOP                  & \textbf{61.62}\% & \textbf{69.97}\% & \textbf{66.09}\% & \textbf{65.55}\% & \textbf{65.82}\%       & \textbf{61.54}\% & \textbf{69.16}\% & \textbf{65.66}\% & \textbf{66.02}\% & \textbf{65.84}\%        \\
\cline{2-12}
                                 & \rule{0pt}{10pt}improvement                & 14.85\% & 15.32\% & 21.22\% & 31.74\% & 27.25\%       & 14.50\% & 14.06\% & 19.76\% & 31.88\% & 26.68\%        \\
\hline
\end{tabular}
\arrayrulecolor{black}
\end{table*}

\subsection{Experimental results}
We report the experimental results from three aspects.

We first consider the performance comparison among different approaches. The performance metrics include $ACC@k$, $macro\text{-}P$, $macro\text{-}R$, and $macro\text{-}F1$. The experimental results are depicted in Table \ref{tab:results}. In the table, we compare the performance of CNNTOP, TULVAE, and Bi-TULER. For Brightkite dataset, we randomly select 100 and 200 users to construct the experimental datasets, which are indicated by Brightkite(100), Brightkite(200), respectively. We do the same on datasets Foursquare and Gowalla. Thus, the experiments are conducted on six datasets. The performance of CNNTOP is shown in bold. The rows just below the CNNTOP performance show the improved values of corresponding metrics. Since different dataset has different inherent properties such as location density, average trajectory length, CNNTOP has different performances on these datasets. Specifically, the performance improvement of $ACC@1$ ranges from 14.85\% to 33.78\%. The improvement of $ACC@5$ ranges from 15.19\% to 28.28\%. The improvement of $macro\text{-}P$ ranges from 16.21\% to 60.62\%. The improvement of $macro\text{-}R$ ranges from 25.19\% to 61.59\%. The improvement of $macro\text{-}F1$ ranges from 21.27\% to 61.11\%. In summary, as to different datasets, CNNTOP far outperforms the baselines at all metrics.

\begin{figure}
  \centering
  \includegraphics[width=0.5\textwidth]{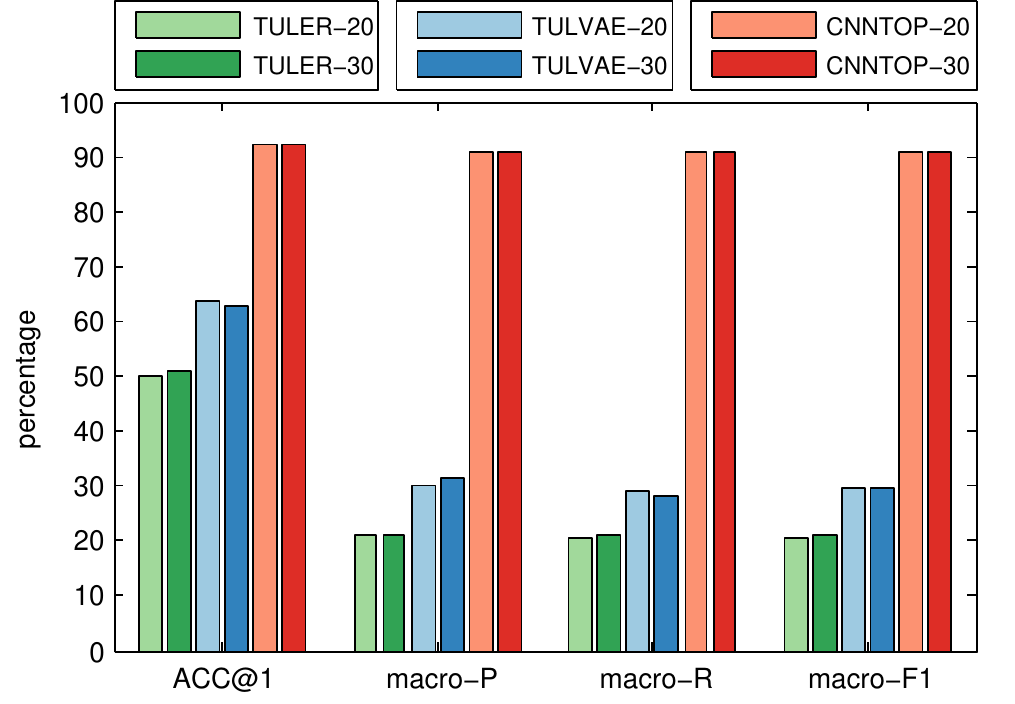}
  \caption{Performance comparison of different approaches with trajectory lengths of 20 and 30 on Brightkite(100).}
  \label{fig:B-bar}
\end{figure}

\begin{figure}
  \centering
  \includegraphics[width=0.5\textwidth]{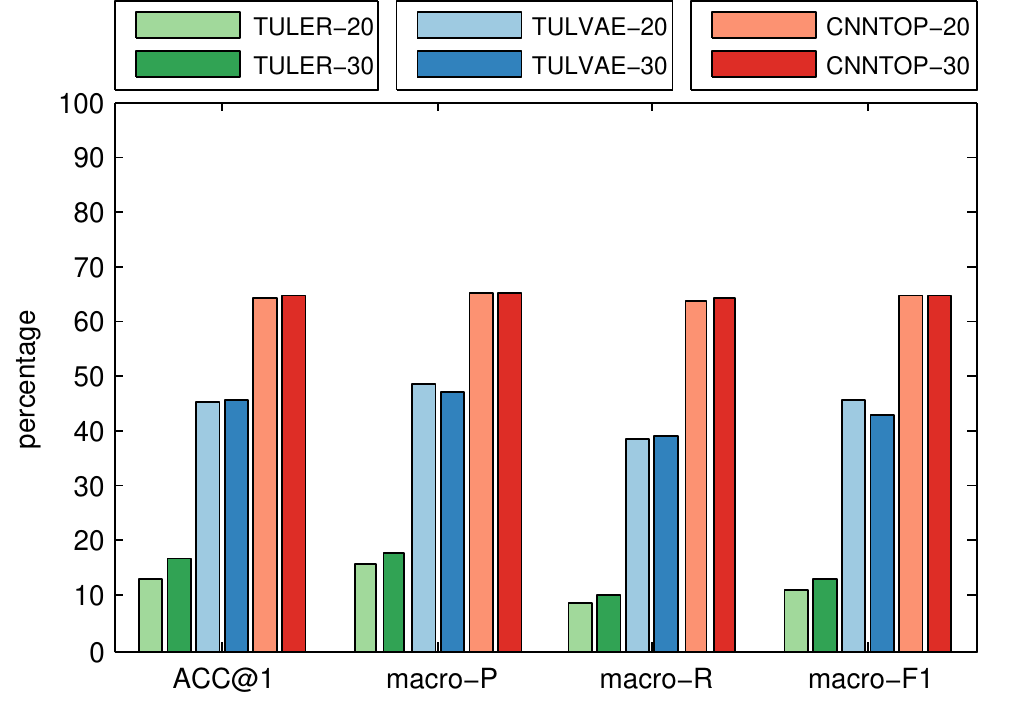}
  \caption{Performance comparison of different approaches with trajectory lengths of 20 and 30 on Foursquare(100).}
  \label{fig:F-bar}
\end{figure}

\begin{figure}
  \centering
  \includegraphics[width=0.5\textwidth]{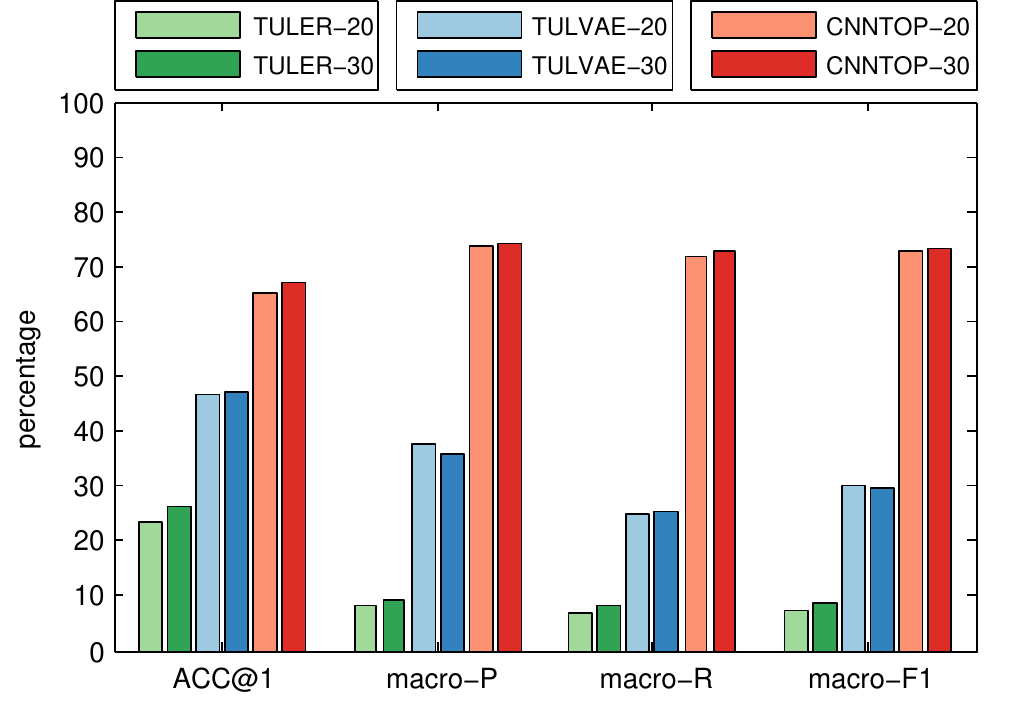}
  \caption{Performance comparison of different approaches with trajectory lengths of 20 and 30 on Gowalla(100).}
  \label{fig:G-bar}
\end{figure}

Then, we evaluate the impact of trajectory length on the prediction performance. If a method has very different performances on different trajectory lengths, its usage scenario is limited.  Therefore, the insensitivity of performance to trajectory length is expected. We evaluate the impact of trajectory length on the prediction performance on Brightkite(100), Foursquare(100), and Gowalla(100), respectively. Trajectory lengths of 20 and 30 are chosen to conduct experiments. The results are depicted in Fig.~\ref{fig:B-bar}, \ref{fig:F-bar}, and \ref{fig:G-bar}, respectively. CNNTOP-20 and CNNTOP-30 are results of CNNTOP corresponding to trajectory length of 20 and 30, respectively. Similarly, Bi-TULER-20 and Bi-TULER-30 are results of Bi-TULER on trajectory length of 20 and 30, respectively. Likewise, TULVAE-20 and TULVAE-30 are results of TULVAE. According to Fig.~\ref{fig:B-bar}, \ref{fig:F-bar}, and \ref{fig:G-bar}, we can see that the performance of CNNTOP is far better than that of Bi-TULER and TULVAE which are two most recently proposed solutions. Specifically, on datasets Brightkite and Foursquare, there are imperceptible differences between CNNTOP-20 and CNNTOP-30. Due to the intrinsic properties of Gowalla dataset, CNNTOP-20 and CNNTOP-30 exhibit different performances. However, the difference is very tiny. According to the above analysis, we can safely conclude that CNNTOP is not sensitive to trajectory length while the baselines have different performances on different settings.

\begin{figure*}
\begin{minipage}[t]{0.5\linewidth}
\centering
\includegraphics[width=1.0\textwidth]{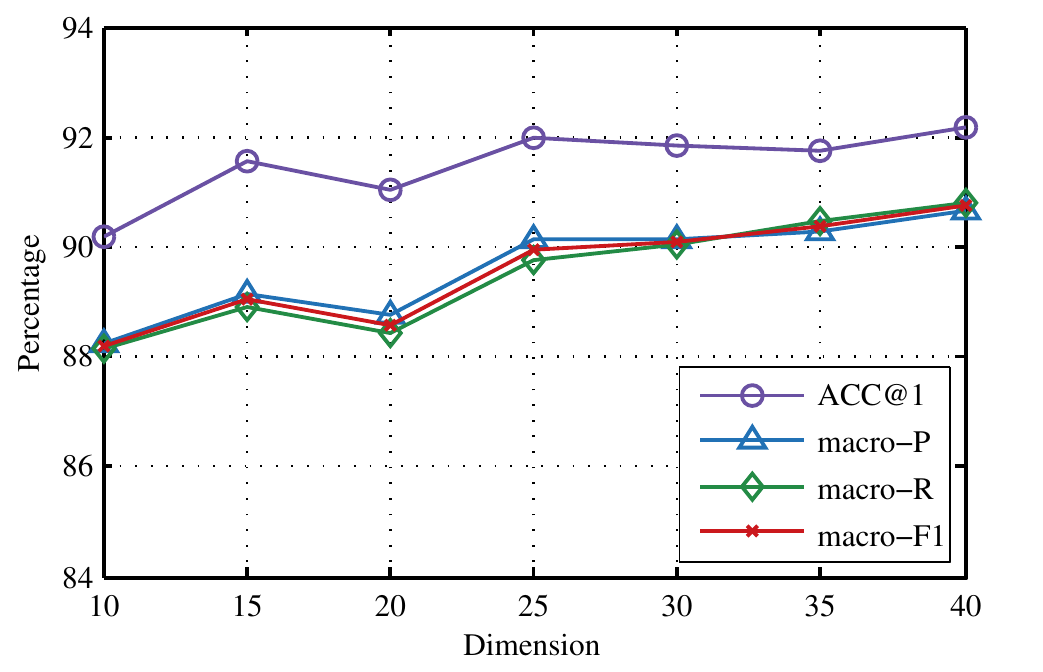}
\caption{Performance impact of POI embedding dimensions on Brightkite with trajectory length 20.}
\label{fig:sens-B-20}
\end{minipage}%
\quad
\begin{minipage}[t]{0.5\linewidth}
\centering
\includegraphics[width=1.0\textwidth]{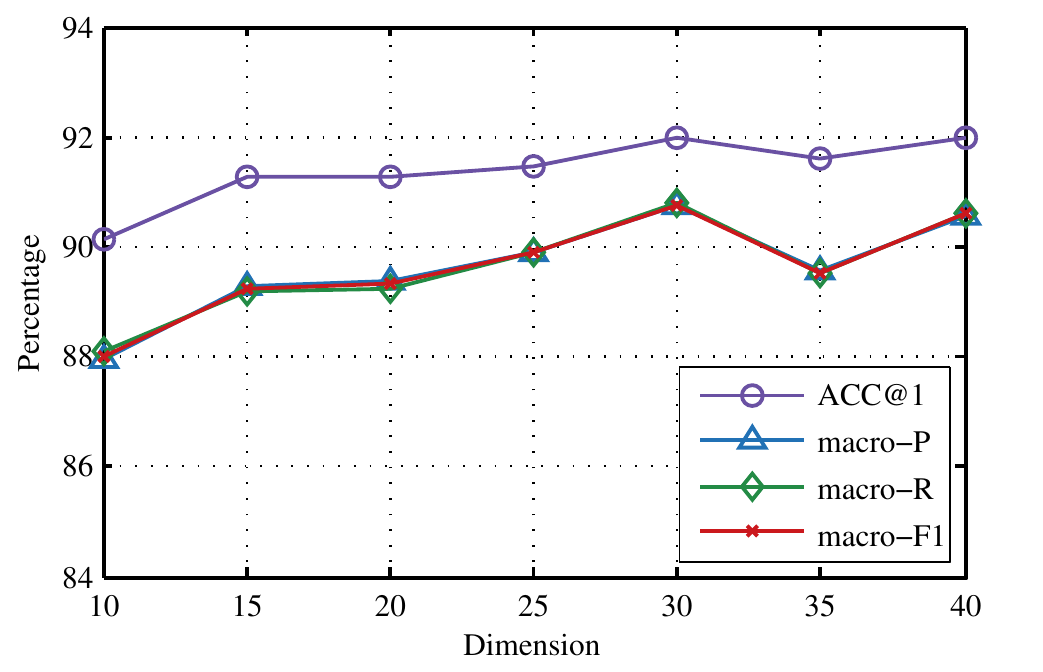}
\caption{Performance impact of POI embedding dimensions on Brightkite with trajectory length 30.}
\label{fig:sens-B-30}
\end{minipage}
\end{figure*}

\begin{figure*}
\begin{minipage}[t]{0.5\linewidth}
\centering
\includegraphics[width=1.0\textwidth]{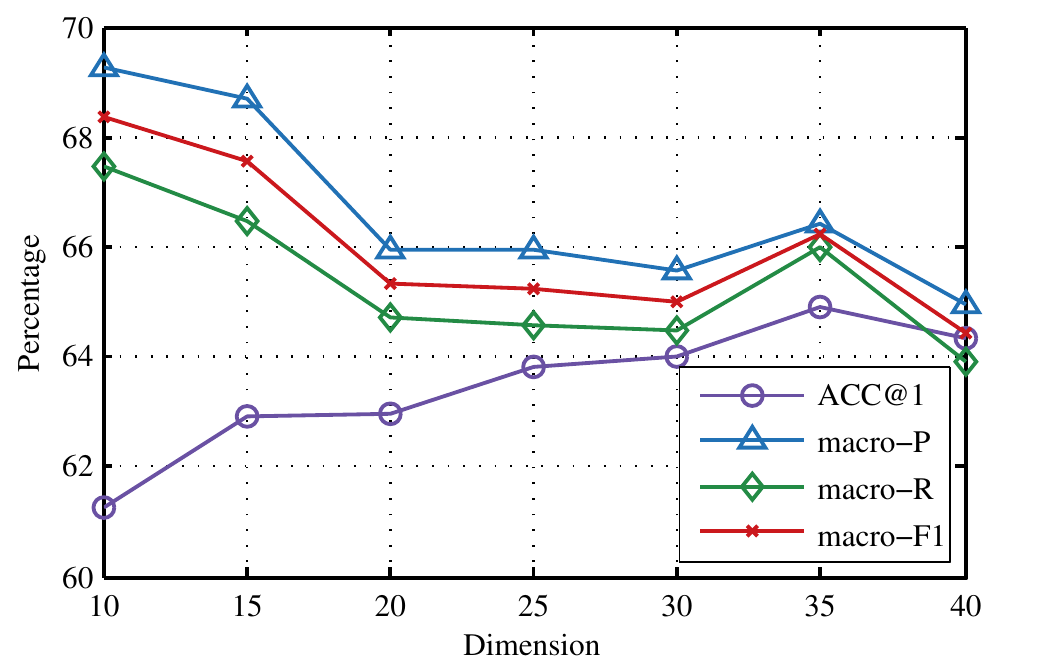}
\caption{Performance impact of POI embedding dimensions on Foursquare with trajectory length 20.}
\label{fig:sens-F-20}
\end{minipage}%
\quad
\begin{minipage}[t]{0.5\linewidth}
\centering
\includegraphics[width=1.0\textwidth]{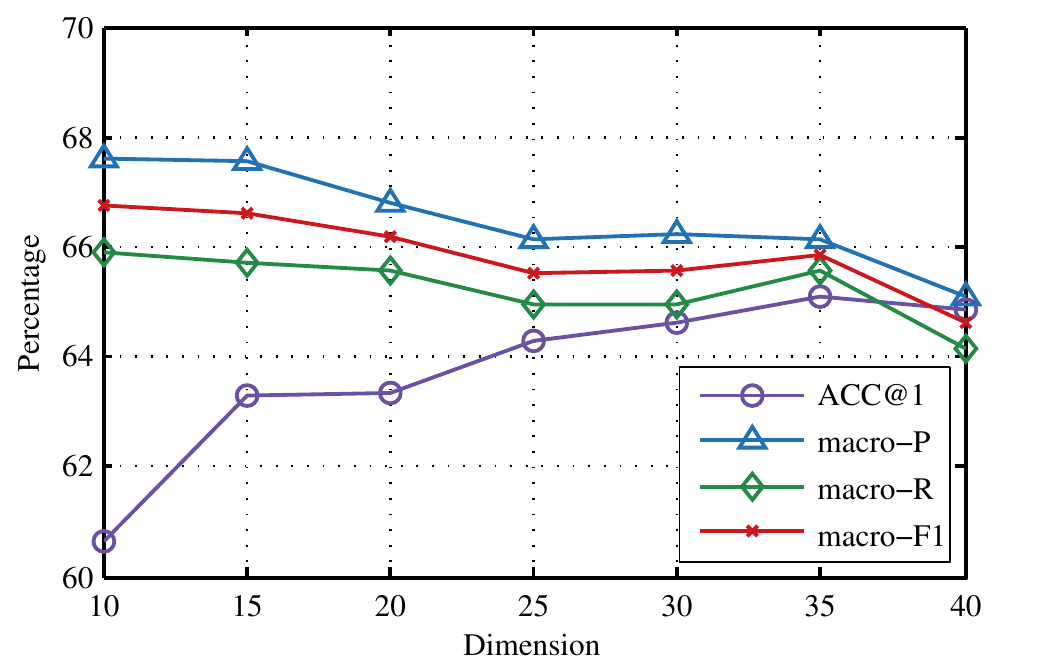}
\caption{Performance impact of POI embedding dimensions on Foursquare with trajectory length 30.}
\label{fig:sens-F-30}
\end{minipage}
\end{figure*}

\begin{figure*}
\begin{minipage}[t]{0.5\linewidth}
\centering
\includegraphics[width=1.0\textwidth]{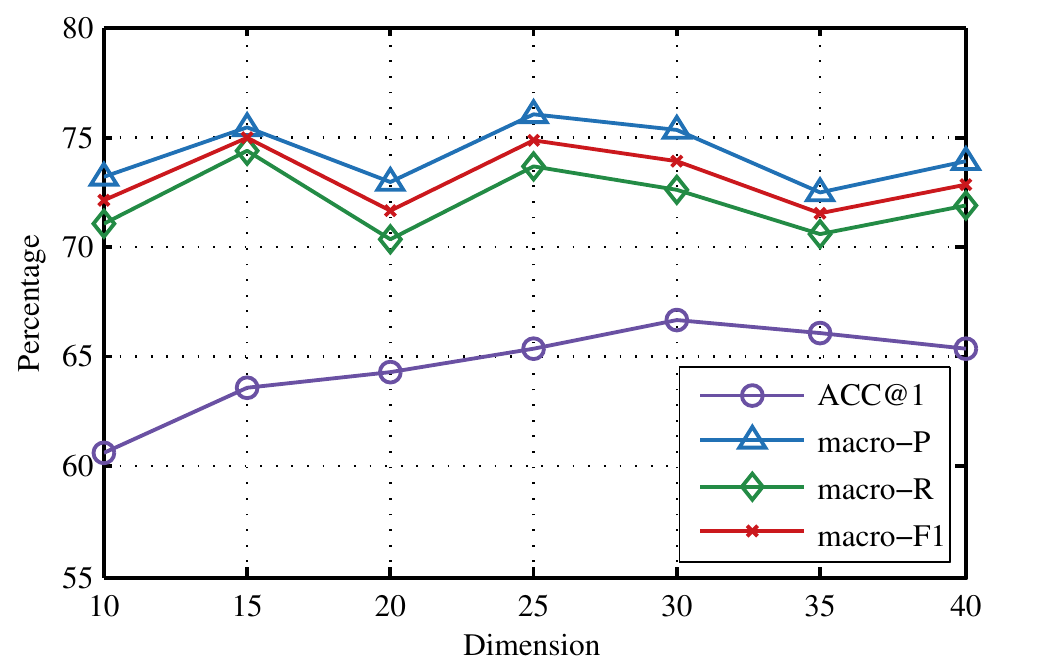}
\caption{Performance impact of POI embedding dimensions on Gowalla with trajectory length 20.}
\label{fig:sens-G-20}
\end{minipage}%
\quad
\begin{minipage}[t]{0.5\linewidth}
\centering
\includegraphics[width=1.0\textwidth]{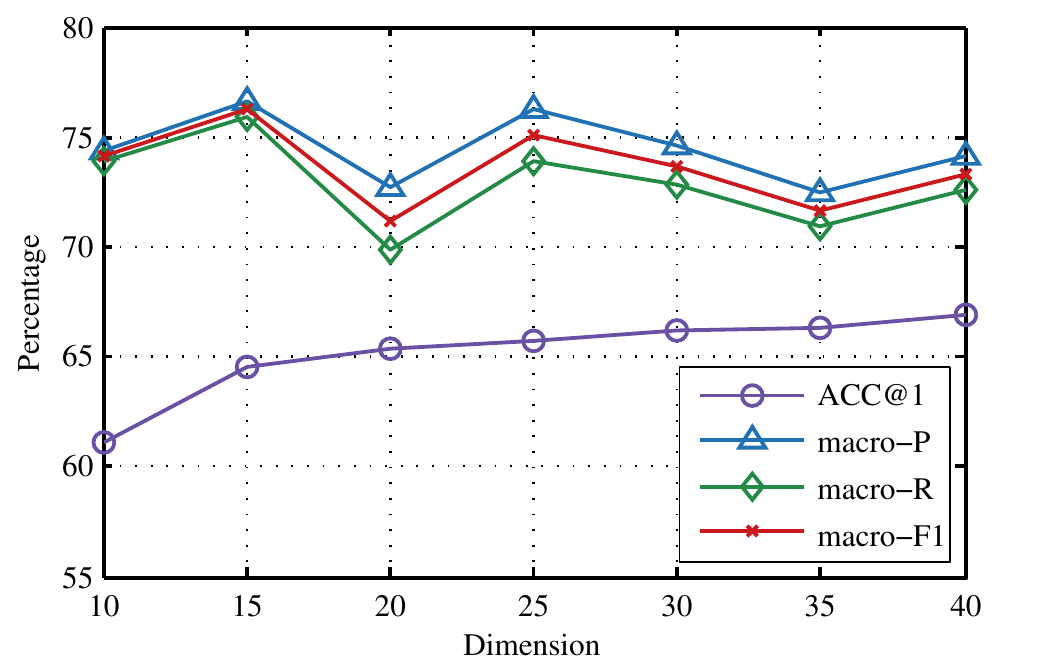}
\caption{Performance impact of POI embedding dimensions on Gowalla with trajectory length 30.}
\label{fig:sens-G-30}
\end{minipage}
\end{figure*}

Finally, we consider the parameter sensitivity of CNNTOP. The parameters considered in the experiments include POI embedding dimensions and trajectory lengths. The results are illustrated in Fig.~\ref{fig:sens-B-20}, \ref{fig:sens-B-30}, \ref{fig:sens-F-20}, \ref{fig:sens-F-30}, \ref{fig:sens-G-20}, and \ref{fig:sens-G-30}. For all the cases, the trends of $ACC@1$ are similar. Specifically, the $ACC@1$ increases as the POI embedding dimension increases. CNNTOP almost achieves the best accuracy when the POI embedding dimension is 30. For dimensions larger than 30, there are no notable fluctuations of accuracy. For other performance metrics, i.e., $macro\text{-}P$, $macro\text{-}R$, and $macro\text{-}F1$, the trends for trajectory lengths of 20 and 30 on the same dataset are also similar. However, there are notable differences on different datasets. Specifically, Fig.~\ref{fig:sens-B-20} shows that the three metrics on trajectory length of 20 decreases a little at the embedding dimension 15 and increases again at the embedding dimension bigger than 20. Figure \ref{fig:sens-B-30} shows that the three metrics on the trajectory length of 30 almost keep increasing from the embedding dimension 10. In contrast with the performance on Brightkite, the above three metrics of CNNTOP on Foursquare are slightly different, which is illustrated in Fig. \ref{fig:sens-F-20} and \ref{fig:sens-F-30}. Generally, the three metrics decrease as the POI embedding dimension increases. Specifically, Fig.~\ref{fig:sens-F-20} shows that the above three performance metrics on trajectory length 20 decreases a little as the POI embedding dimension increases from 10. From dimension 20 to 30, the three metrics keep stable. After that, the three metrics increase and decrease again at dimension 35. However, the ranges of fluctuations are all less than five. Figure \ref{fig:sens-F-30} shows that CNNTOP with trajectory length of 30 has similar performance on the dataset Foursquare. When it comes to CNNTOP on the Gowalla dataset, the performances are very similar, which are illustrated in Fig.~\ref{fig:sens-G-20} and Fig.~\ref{fig:sens-G-30}. The trends of accuracy with different POI embedding dimensions are similar to that of Brightkite and Foursquare. As to trajectory lengths of 20 and 30, the metrics $macro\text{-}P$, $macro\text{-}R$, and $macro\text{-}F1$ increase from POI embedding size of 10. When the embedding dimension is 25, all the performance metrics almost achieve the best and there are some small fluctuations after that.

According to the above analysis, the trends of accuracy with different settings are very similar. Due to the intrinsic properties of different datasets, such as POI density, average trajectory length, the trends of $macro\text{-}P$, $macro\text{-}R$, and $macro\text{-}F1$ are slightly different. However, the general fluctuations are very small. Therefore, the accuracy of CNNTOP is not sensitive to different datasets while the other three metrics are slightly different on different datasets. As to the same dataset, the impact of trajectory length on performance is limited. In short, for practical usage, we recommend setting the POI dimension as 30 to achieve good performance.

\section{Conclusion and future work}
\label{Conclusion}
In this work, we propose CNNTOP to address the trajectory owner prediction. CNNTOP includes two parts. The first part maps POIs to low-dimension real value vectors. Based on the POI vectors, the matrix representations of trajectories are constructed. The second part is a CNN-based classifier that maps the matrix representations of trajectories to probability distributions of trajectory owners. Extensive experiments demonstrate that CNNTOP outperforms existing solutions. In the future, we will build an end-to-end model to address this problem since task-specific POI representation can efficiently boost performance.

\section*{Acknowledgment}
This work was supported by the National Science Foundation of China under grant No.61272527.

\bibliographystyle{IEEEtran}
\bibliography{IEEEabrv,trajectory-cnn-ref-luo}

\end{document}